%% file: main.tex
\documentclass[final]{cvpr}

\usepackage{times}
\usepackage{epsfig}
\usepackage{graphicx}
\usepackage{amsmath}
\usepackage{amssymb}
\usepackage{multirow}
\usepackage{adjustbox}

\usepackage{enumitem}
\usepackage{amsthm}
\usepackage[ruled,vlined,linesnumbered]{algorithm2e}

\usepackage[
  separate-uncertainty = true,
  multi-part-units = repeat
]{siunitx}

\usepackage[pagebackref=true,breaklinks=true,colorlinks,bookmarks=false]{hyperref}

\def\cvprPaperID{} 
\def\confYear{CVPR 2021}

\begin{document}

\title{Towards Solving Multimodal Comprehension}


\author{

Pritish Sahu$^{2}$\thanks{†This work was done as an intern at SRI International.} \quad Karan Sikka$^{1}$ \quad Ajay Divakaran$^{1}$\\
$^{1}$SRI International \\
$^{2}$Rutgers University \\
\\
}

\def\algorithmautorefname{Algorithm}
\def\figureautorefname{Figure}
\def\tableautorefname{Table}
\def\equationautorefname{Eq.}
\def\sectionautorefname{Section}

\maketitle

\def\ie{\textit{i.e.}}
\def\eg{\textit{e.g.}}

\definecolor{redcol}{rgb}{1, 0, 0}
\definecolor{bluecol}{rgb}{0, 0, 1}
\newcommand{\red}[1]{\textcolor{redcol}{#1}} 
\newcommand{\blue}[1]{\textcolor{bluecol}{#1}} 
\renewcommand{\paragraph}[1]{\smallskip\noindent{\bf{#1}}}
\newcommand{\todo}[1]{\red{TODO: {#1}}}
\newcommand{\colons}[1]{``{#1}''}

\def\algorithmautorefname{Algorithm}
\def\figureautorefname{Figure}
\def\tableautorefname{Table}
\def\equationautorefname{Eq.}
\def\sectionautorefname{Section}

\input{abstract}

\input{intro}

\input{related_work}

\input{approach}

\input{exp}

\input{conclusion}

{\small
\bibliographystyle{ieee_fullname}
\bibliography{egbib}
}


\end{document}

%% file: abstract.tex

\begin{abstract}
    This paper targets the problem of procedural multimodal machine comprehension (\textit{M3C}). This task requires an AI to comprehend given steps of multimodal instructions and then answer questions. 
    Compared to vanilla machine comprehension tasks where an AI is required only to understand a textual input, procedural M3C is more challenging as the AI needs to comprehend both the temporal and causal factors along with multimodal inputs.
    Recently Yagcioglu \etal \cite{yagcioglu2018recipeqa} introduced RecipeQA dataset to evaluate M3C. 
    Our first contribution is the introduction of two new M3C datasets-- \textit{WoodworkQA} and \textit{DecorationQA} with $16K$ and $10K$ instructional procedures, respectively. 
    We then evaluate M3C using a textual cloze style question-answering task and  
    highlight an inherent bias in the question answer generation method from \cite{yagcioglu2018recipeqa} that enables a naive baseline to cheat by learning from only answer choices.
    This naive baseline performs similar to a popular method used in question answering-- Impatient Reader \cite{hermann2015teaching} that uses attention over both the context and the query.
    
    We hypothesized that this naturally occurring bias present in the dataset affects even the best performing model. We verify our proposed hypothesis and  propose an algorithm capable of modifying the given dataset to remove the bias elements. Finally, we report our performance on the debiased dataset with several strong baselines. We observe that the performance of all methods falls by a margin of $8\% - 16\%$ after correcting for the bias. We hope these datasets and the analysis will provide valuable benchmarks and encourage further research in this area. 
\end{abstract}

%% file: intro.tex
\section{Introduction}
\label{sec:intro}





A key focus in machine learning has been to create models that can comprehend any given document and then answer questions that require contextual, temporal, and cross-modal reasoning \cite{rajpurkar2016squad, antol2015vqa, richardson2013mctest, kembhavi2016diagram}. 
The importance of this task has led to its popularity in both NLP and Computer vision as the Reading Comprehension (RC) and Visual Question Answering (VQA) tasks, respectively.

In the RC task, a model is typically required to answer a question conditioned on a given document (referred to as \textit{context}). The answer could be located directly in the text \cite{rajpurkar2016squad} or could be inferred through some reasoning such as temporal \cite{liu2019machine}, or common-sense \cite{huang2019cosmos}. Research in this area has been propelled by multiple datasets from different domains and of varying complexity \cite{rajpurkar2016squad,hewlett2016wikireading,kovcisky2018narrativeqa, liu2019machine, hill2016goldilocks, hermann2015teaching, trischler2017newsqa}.
Visual Question Answering (VQA) \cite{antol2015vqa,goyal2017something} is also a type of RC task, where a model is asked to answer a text-based question about a visual context. Similar to RC, VQA requires different kinds of reasoning such as spatial \cite{johnson2017clevr}, common-sense \cite{zellers2019recognition}, contextual \cite{antol2015vqa}. 
Recently, the above\-mentioned tasks have been extended to solve a new task known as multimodal machine comprehension (M3C). Here the context contains both textual and visual information \cite{tapaswi2016movieqa, iyyer2017amazing, kembhavi2016diagram, kahou2017figureqa, kembhavi2017you}. In this case, a model will also have to establish the correspondence between the two modalities for question answering.

In this work, we focus on procedural M3C where the context is a series of procedural or How-To steps describing the entire process of preparing or constructing something (see \autoref{fig:woodwork_sample}). Each such step is described by multiple images, text, or both. We focus on cloze style question answering where a model that learns to comprehend multi-modal sources of information can fill the missing step in the question.
Answering such questions would depend on jointly modeling both textual and visual content from each step while simultaneously understanding the temporal evolution of the steps. 
Our work builds upon \cite{yagcioglu2018recipeqa} that introduced the RecipeQA dataset for the procedural understanding of food recipes. The lack of procedural M3C benchmarks motivated us to introduce two new datasets-- \textit{WoodworkQA} and \textit{DecorationQA}, to evaluate performance on procedures relating to woodwork and decorative items respectively. We chose Woodwork and Decoration due to the consistency of user projects across all subcategories and the availability of a large number of instructional procedures. These datasets were collected from the website Instructables\footnote{https://www.instructables.com/} and  include natural images and text written by real users.
We then identify a limitation with the question generation method for the cloze task used in \cite{yagcioglu2018recipeqa}, enabling a naive baseline, that looks only from answer choices without the context, to cheat and achieve results closer to ImpatientReader that uses context, question and choices. That would thus prevent unbiased evaluation of new methods on this problem. 
We address this problem through an algorithm for automated question answer generation by uniformly distributing the answer choices. We finally evaluate several baseline models based on LSTM and BERT on these datasets and discuss these analyses. 

\begin{figure*}[htbp!]
    \begin{center}
       \includegraphics[width=0.9\linewidth]{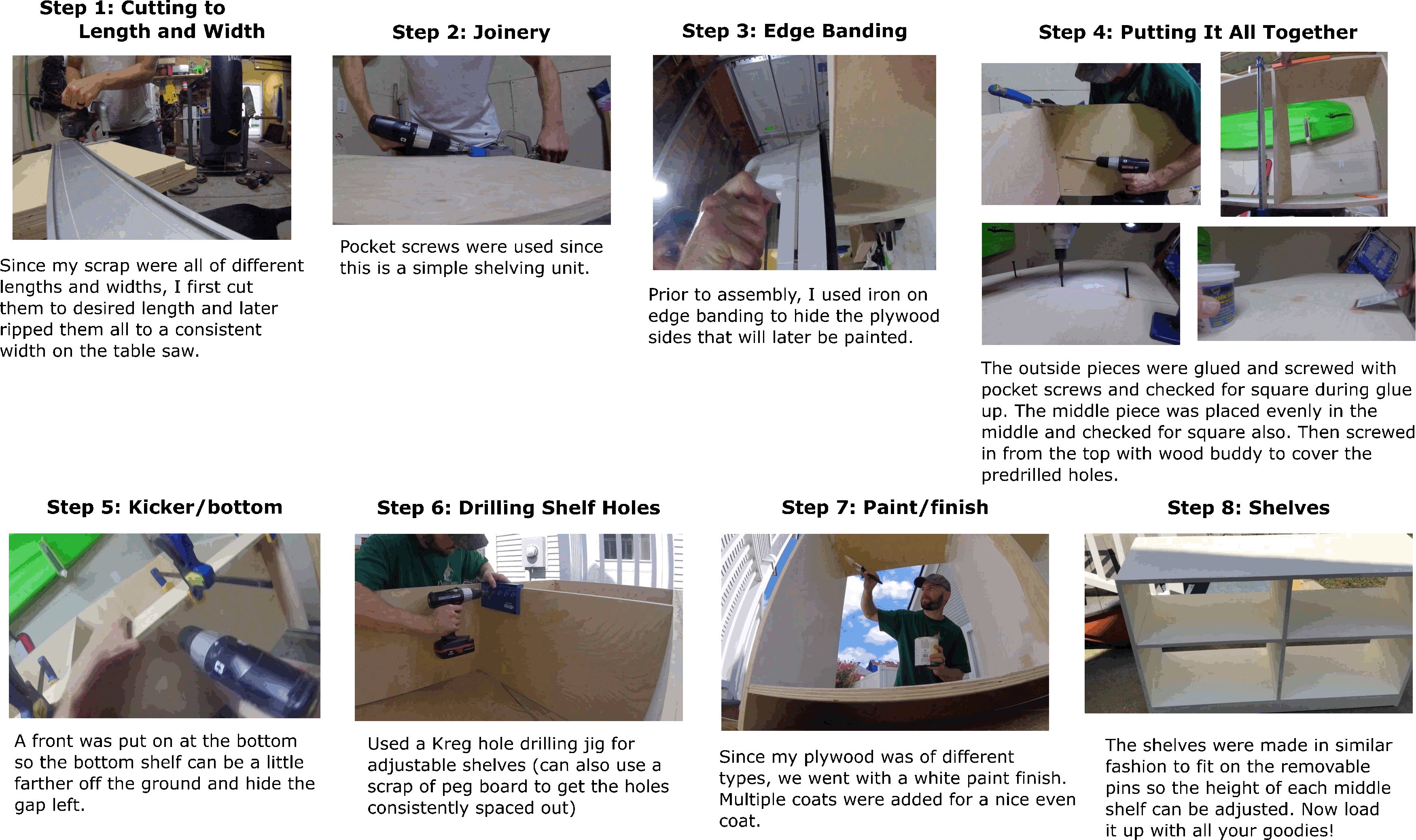}
    \end{center}
    \label{fig:woodwork_original}
    \caption{Woodwork: Procedural description on how to make a \colons{Plywood Bookshelf} from scratch, as described in Instructables. Each step contains natural language and images describing the process.}
 \end{figure*}

Overall, the contributions proposed in this paper are as follows:
\begin{itemize}
    \item We study and evaluate the task of multimodal machine comprehension that consists of answering the question presented in cloze style based on a given procedure in multiple modalities. 
    \item We highlight an inherent drawback in the way datasets have been used for evaluation in that it allows a naive ML model to cheat by using only the choices/answers without access to the pertinent questions. This naive model performance gets closer to one of the popular method used in question answering -- i.e. ImpatientReader \cite{hermann2015teaching}.
    \item We propose an automated method to build training and evaluation datasets from these procedures that are free from the above mentioned bias.
    \item We introduce two new datasets and evaluate several baselines on three datasets including RecipeQA. The dataset will soon be available for downloaded from \url{sahupritish.github.io/multimodal_comprehesion/}.
    \item We believe our work will provide sound and robust benchmarks for further research in this area.
\end{itemize}

%% file: related_work.tex
\section{Related Works}

\begin{figure*}[t]
    \begin{center}
       \includegraphics[width=0.9\linewidth]{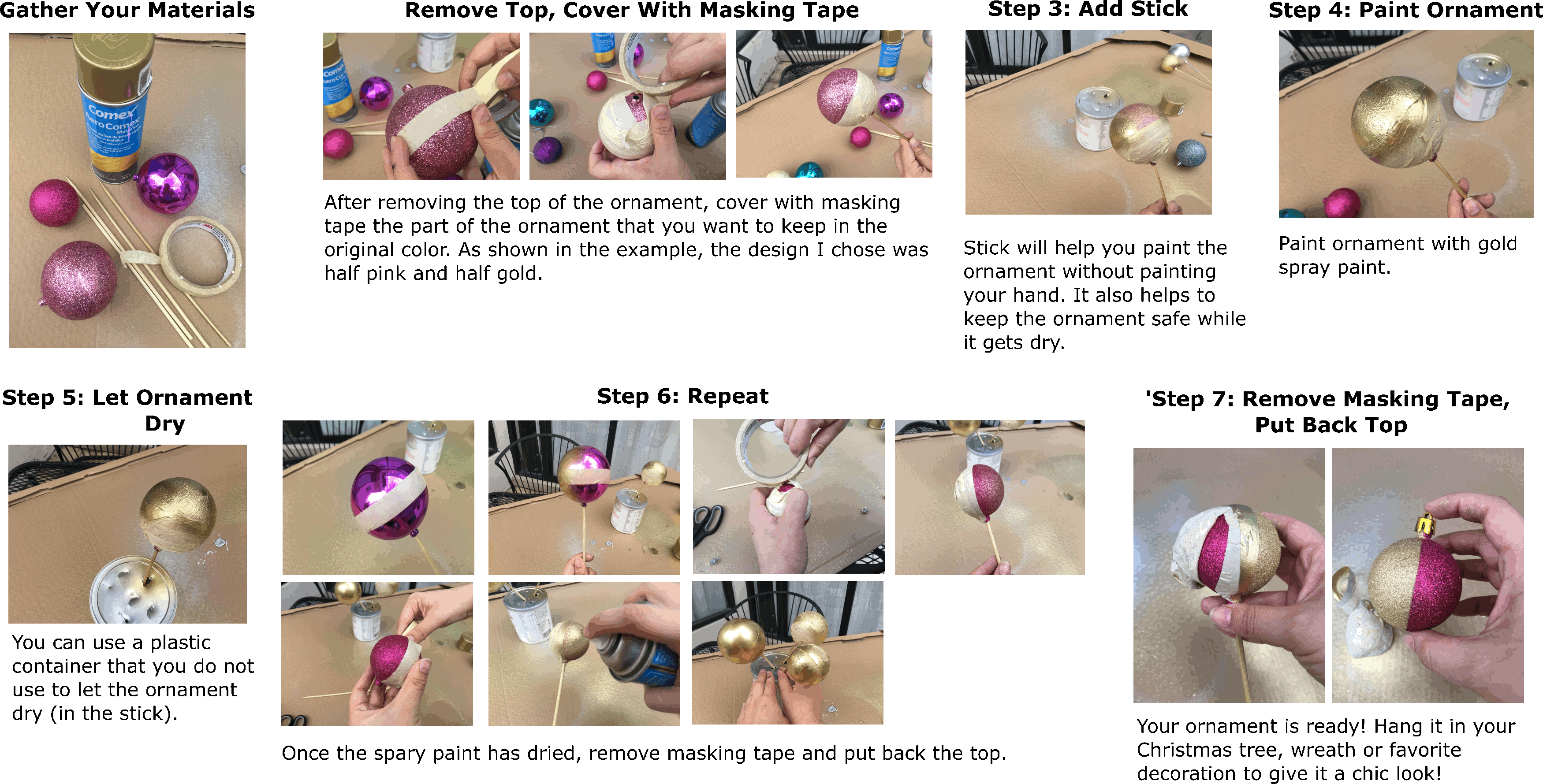}
    \end{center}
    \label{fig:decoration_original}
    \caption{Decoration: Procedural description on how to make a \colons{Rose-Golden} colored Christmas ornament from scratch, as described in Instructables. Each step contains natural language and images describing the process.}
 \end{figure*}


\textbf{Visual Question Answering (VQA)}  is a joint learning task in vision and language where the question-answer are in the form of natural language, and the context is a natural image. Over the recent years, VQA has been studied extensively due to the emergence of large-scale datasets such as MSCOCO \cite{lin2014microsoft}, Flickr30K \cite{young2014image}, and Abstract Scenes \cite{zitnick2014adopting}. 
One of the earliest datasets introduced was DAQUAR, with a few thousand question-answer pairs \cite{malinowski2014multi}. Since then, larger datasets such as COCO-QA \cite{ren2015exploring}, FMIQA \cite{gao2015you}, Visual Madlibs \cite{yu2015visual}, and VQA \cite{antol2015vqa} have been released.  
In contrast to VQA, where the context is visual, the context in the proposed datasets contains both natural images and language. The context is also temporal in nature and provides a step-wise description of a wood crafting or a decoration related projects.
Some of the early approaches in VQA typically used RNN-CNN based networks that first encodes both the question and the context, and combines them to answer the question asked \cite{antol2015vqa,malinowski2015ask}. With the advent of attention-based mechanisms, many works introduced an attention module that involves attending to parts of an image using the entire question or parts of the question \cite{xiong2016dynamic,xu2016ask,zhu2016visual7w}. \cite{lu2016hierarchical} employed cross attention that involved attending to both the context and the question. The above approaches to VQA would suffer in the proposed datasets in the following way. First, VQA model's learns on question-answer pair whereas in M3C, the model has to jointly understand the multimodal data at each step, understand the temporal evolution and use the tuple (context, question, choice) to predict the correct answer. Second, VQA requires the model to possess common sense to answer many questions, which would be possible with a model trained on a large corpus of data or an additional knowledge base model, while M3C is procedural and confines the required knowledge from beginning to end in the context.

\textbf{Machine Reading Comprehension}
(RC) is another form of question-answering task in NLP. One of the common ways to set up an RC task is via cloze styled questions where a model learns to read and comprehend a given text passage and answers question about it. RC has been a growing field of research due to the availability of several RC datasets introduced over the years, such as CNN/Daily Mail \cite{hermann2015teaching},  Children's Book Test \cite{hill2016goldilocks},  MCTest \cite{richardson2013mctest},  Stanford QuestionAnswering (SQuAD) \cite{rajpurkar2016squad}, that are either sourced by crowdworkers or via automation. Some of the popular approaches in RC are based on variations of attention mechanism \cite{hermann2015teaching,chen2016thorough}, memory networks \cite{weston2014memory,xiong2016dynamic}.

The two categories of datasets discussed above measure different aspects of the comprehension and reasoning task. In contrast, DecorationQA and WoodworkQA can be used to measure confined knowledge and reasoning. Following are some of the recent works that share similarities to our datasets, such as COMICS \cite{iyyer2017amazing},  TQA \cite{kembhavi2017you}, MoviesQA \cite{tapaswi2016movieqa}. These datasets incorporate multimodality in their context. 
COMICS \cite{iyyer2017amazing} dataset provides cloze styles question answers using extracts from comic books that contain drawings and text blobs describing a scene.  MoviesQA \cite{tapaswi2016movieqa} dataset uses movie clips, subtitles, and snapshots to prepare the M3C dataset. Similarly, TQA \cite{kembhavi2017you} uses middle school science lessons, especially the diagrams in the books with images and text, to provide multimodal context. In contrast, our datasets-- WoodworkQA and DecorationQA  the context contains the entire procedure as well the natural image and text are collected from anonymous users in unconstrained environments. 

Finally, our new additional M3C dataset provides the researchers with more benchmarks to support proposed models. In our work, we also noticed the presence of inherent bias in the data, which could be caused by the form template every user needs to fill describing the entire process or the algorithm used by \cite{yagcioglu2018recipeqa} in preparation of the dataset.

%% file: approach.tex
\section{Multi Modal Machine Comprehension Dataset}

\begin{table*}[!htbp]
   \centering
   \begin{tabular}{lccccc}
   \hline
                              & \multicolumn{2}{c}{WoodworkQA}                       & \multicolumn{1}{c}{} & \multicolumn{2}{c}{DecorationQA}                      \\ \cline{2-6} 
   Model                      & Train                     & Val                      &                      & Train                     & Val                       \\ \hline
   \# projects                & $14625$   & $1807$     &    & $9219$   & $1139$  \\
   avg. \# of steps           & $8.01$  & $8.06$  &  & $6.82$  & $6.59$  \\
   avg. \# of tokens (titles) &  $27.03$  &  $28.44$   &   &   $22.44$   &   $21.77$    \\
   avg. \# of tokens (descr.) & $781.27$      &  $804.87$    & & $527.53$ & $543.20$   \\
   avg. \# of images          & $9.98$      & $10.8$     &  & $7.59$  & $12.36$ \\
   \# textual cloze           & $23698$ & $2656$ &  & $12793$ & $1414$  \\ \hline
   \\
   \end{tabular}
   \caption{WoodworkQA and DecorationQA statistics}
    \label{table:statistics}
\end{table*}

We introduce two new multimodal datasets, \textit{WoodworkQA} and \textit{DecorationQA} to evaluate reasoning capabilities 
required to answer a question pertaining to real-world crafting procedure. 
These datasets are prepared to provide a step-wise temporal coherent story from collecting items necessary for the project to the final step of completing the project. 
The arbitrary number of steps in the context of our M3C data provides a complete procedural story, from collecting the materials required for the project to the completion of the project.

Other introduced  M3C works such as, \cite{tapaswi2016movieqa,iyyer2017amazing,kembhavi2017you} present significant shortcomings for any model because the context contains only partial knowledge to the entire story, such as extracts from comics or clips and subtitles from movies which fails to deliver information on the events that led to the current scenario. In contrast, to answer textual cloze style questions-answering in procedural M3C, a reasonable model trained should comprehend the entire instructional flow of the event to fill in the missing step in the  question list. To this end, any model which works on incomplete data instances learns to predict without ever understating the entire storyline.

WoodworkQA consists of approximately 16K different woodwork procedures collected from over four different wood crafting projects--\colons{Furniture}, \colons{Shelves}, \colons{Workbenches} and \colons{Woodwork}. Similarly, DecorationQA consists of approximately 10K different decorating procedures collected from four decoration related categories, which are \colons{Christmas}, \colons{Halloween}, \colons{Holidays}, \colons{Decorating}. We prepare over 26K and 14K textual cloze type questions-answers for \textit{WoodworkQA} and \textit{DecorationQA}, respectively. Please refer to \autoref{table:statistics} for details on the statistics for each dataset.

Each individual woodwoork/decoration item contains a \colons{Project Title}, \colons{Description} i.e. step wise illustration of the project from start to end, where each step comprises of (i) title of step, (ii) description of the step and (iii) natural images illustrating that step. For supervised learning purpose, we have prepared the dataset similar to \cite{yagcioglu2018recipeqa}, where the tasks explore four types of multimodal aspect of the dataset namely Textual Cloze, Visual Cloze, Visual Coherency, Visual Ordering. We only focus on text cloze style questions in this work which is described in \autoref{textualclozeref}. Below, we describe the data collection procedure and the question-answer preparation methodology for \textit{WoodworkQA} and \textit{DecorationQA} dataset.

\subsection{Dataset Collection}
\textit{WoodworkQA} and \textit{DecorationQA} are collected from the website Instructables\footnote{https://www.instructables.com/}. The website is an online platform hosting \colons{How-To} in a step--by--step procedure on various projects such as circuit design, decoration, cooking. It is an online community where users share the process they followed to complete a project in a step-by-step manner. 

We describe below our methodology involved in collecting the datasets. We inspected the entire Instructables website to select \colons{How-To} categories with sufficient examples and subcategories to make the dataset challenging. 

Out of six main categories in Instructables, we chose \colons{Workshop} and \colons{Living}. Under \colons{Workshop}, there are $25$ subcategories and after carefully exploring we found subcategories \colons{Furniture}, \colons{Shelves}, \colons{Woodwork}, and \colons{Workbenches} to successfully meet the criteria mentioned above. 

We followed a similar procedure in \colons{Living} to select subcategories \colons{Christmas}, \colons{Decorating}, \colons{Halloween}, and \colons{Holiday}. In order to obtain high quality data that includes clear instructions with high-quality descriptive images, we employed a set of heuristics. Firstly, we prepared a script to automate the scraping of data in each subcategory by sorting them in the decreasing order of popularity among readers. Out of 16K and 10K unique woodwork and decoration projects, we further filtered out projects containing non-English text using \cite{lui2012langid}. To have multi-modal data, we also removed projects that only had images or videos and no textual content.  Currently, we do not include video in the step wise data description as there are very few projects that provide video as a form of modality. Finally, the textual content of the remaining projects was filtered to remove non-ASCII characters.

\begin{figure*}[t]
   \begin{center}
      \includegraphics[width=0.9\linewidth]{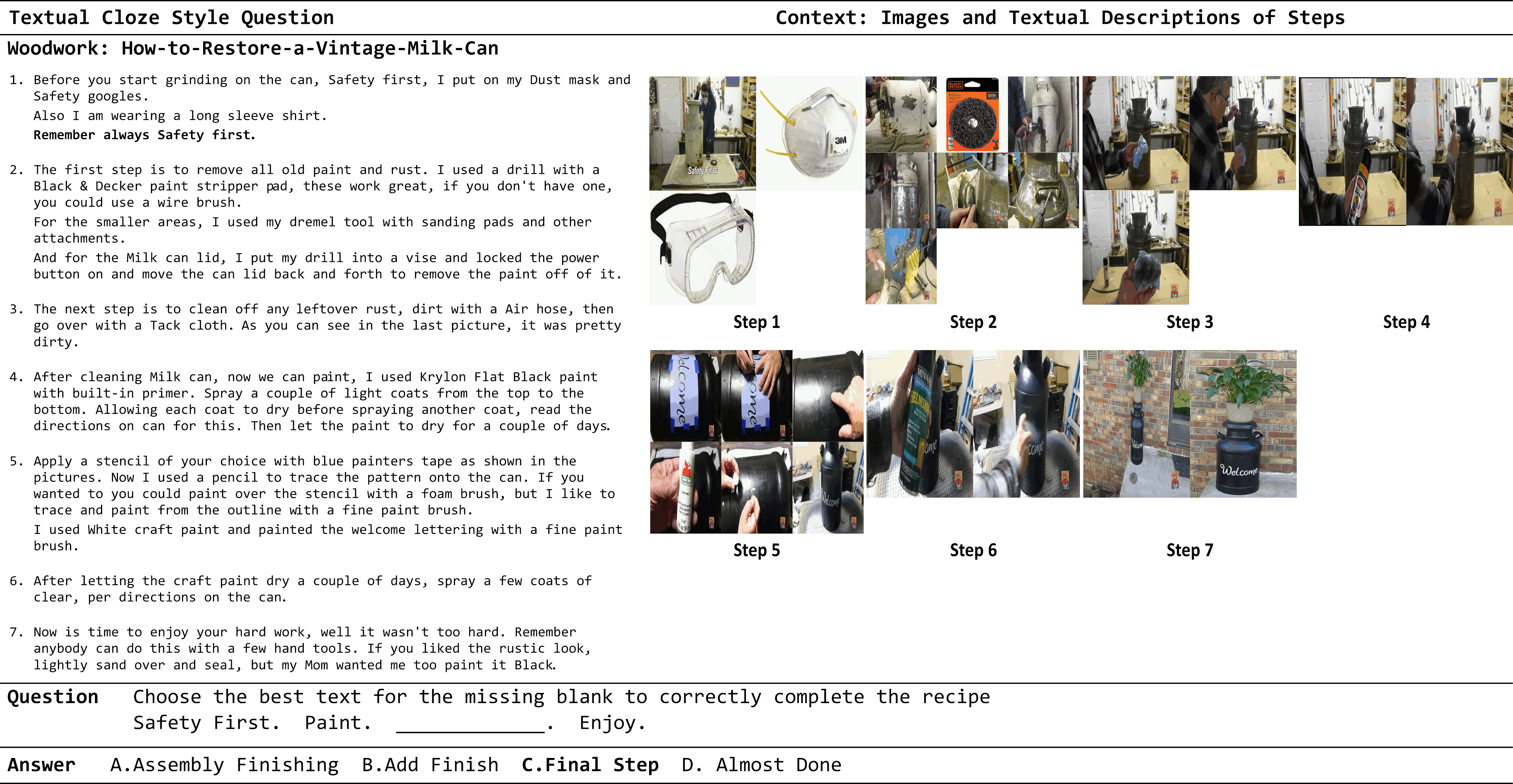}
   \end{center}
   \caption{A sample textual cloze style question from WoodworkQA. The question consists of context, question and answer triplet. The context comprises of step wise description in the form of natural language and images for crafting a  wood related project. The question is formed by using the step titles in a temporal order where the missing title is the bold text from the answer section.}
   \label{fig:woodwork_sample}
\end{figure*}

\subsection{Textual Cloze Style Questions} \label{textualclozeref}
\begin{figure*}[t]
    \begin{center}
       \includegraphics[width=0.9\linewidth]{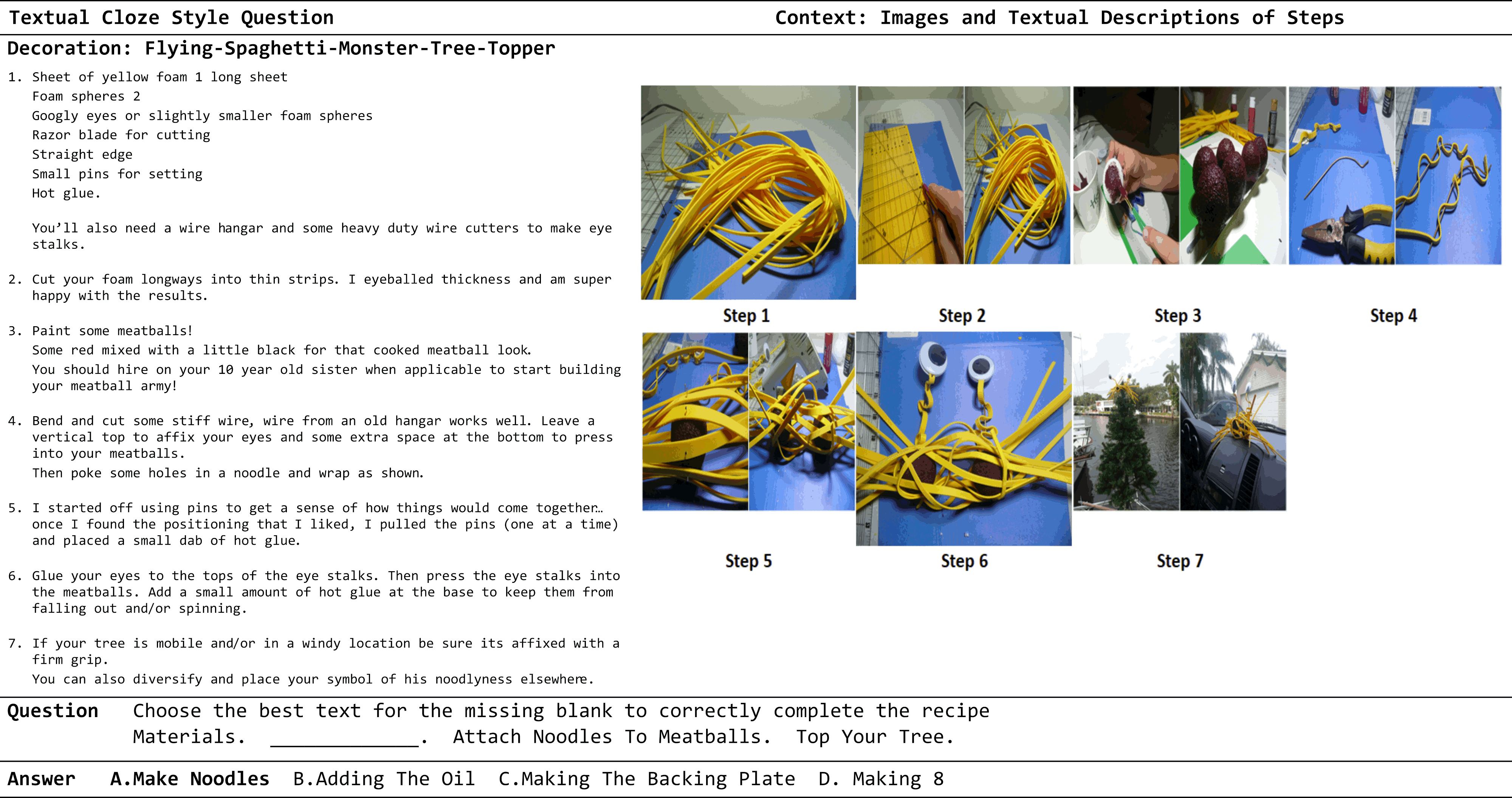}
    \end{center}
    \caption{A sample textual cloze style question from DecorationQA. The question consists of context, question and answer triplet. The context comprises of step wise description in the form of natural language and images for crafting a  decoration related project. The question is formed by using the step titles in a temporal order where the missing title is the bold text from the answer section.}
    \label{fig:decoration_sample}
 \end{figure*}

The template used to construct textual cloze questions consists of context with all the step-wise data except the step titles. The question is prepared by arbitrarily selecting some step titles sequentially and then removing a random title from the candidate list. 
The answer choices consist of the correct step title and incorrect step titles that do not fit to the temporal order of the candidate list (see \autoref{fig:decoration_sample} and \autoref{fig:woodwork_sample} for the visual explanation). For our experiments, we fixed the size of the candidate list and choice list to four. 
In this task, the model needs to pick the correct missing step from the list of choices based on the given multimodal context.  
We generated question-answer pairs for each woodwork/decoration task by randomly removing one title from the candidate list and placing it at a random location in the choice list. 
We get the incorrect choices through simple heuristics mentioned in \cite{yagcioglu2018recipeqa}. We used a NLP model such as Word2Vec or RoBERTA to prepare the textual embeddings for the titles. From these embeddings, we use k nearest neighbor \cite{2020SciPy-NMeth} on the embedding of the correct answer in order to get the nearest titles with their distance values for the correct step title. 
In the next step, the titles that are too close are filtered via an adaptive neighborhood approach. We removed the
step titles with a distance lower than the mean distance over
the titles we received from our k nearest neighbor (k=100) for our adaptive neighborhood. The above step is required since close titles can correctly replace the correct title in the sequence leading to multiple correct answers.  
We then randomly select three titles from the list as incorrect choices. Finally, we remove indicators in the step title in the form of step numbers to prevent a bias in the data that will allow a model to memorize the temporal order. 
We also filtered out the recipes that contained less than $3$ or more than $25$ steps.

\subsection{Bias Issue and Proposed Solution}

During our investigation of this procedural M3C datasets, we discovered an inherent bias in the data. 
This bias leaks information about the target label in an undesired way and thus enables a model to cheat without learning anything useful.
We show this bias in the data by training a baseline that only uses answer choices. 
This baseline is similar to a student who would try to answer questions about a passage without reading it and instead uses patterns from the answer choices e.g. filtering out choices that have a low likelihood of being correct. 
We observe from \autoref{table:biased} that the performance of this baseline (\textbf{hasty student (choice only)}) is significantly greater than random choice and comes close to Impatient Reader. 
This hidden pattern will make evaluating models harder as they would try to exploit this bias instead of understanding the context.
One of the key reasons responsible for this bias is the question preparation method used in RecipeQA \cite{yagcioglu2018recipeqa}, which suffers from distributional bias of the choices. 
For example, most of the beginning and final step titles contain words such as \colons{Ingredients} and \colons{Enjoy}, so if the answer to any question is either the start or the final step, the model will learn to prefer titles containing these words. 
Another issue is to include recipes with a small step count (e.g. $\leq 6$) since the list of question titles will contain three titles, and then the model has to pick the missing one from the remaining three, which is easier for the model to memorize.

We propose \autoref{algorithm} that tries to distribute uniformly the answer choices and thus making  
it more difficult to answer question by only using answer choices. 
The algorithm takes as input the biased train ($\mathcal{D}_{train}$) and validation ($\mathcal{D}_{val}$) dataset and an NLP model such as Word2Vec (trained on the dataset) or RoBERTA (pre-trained on a large corpus). 
We collect all the step titles from ($\mathcal{D}_{train}$, $\mathcal{D}_{val}$) and encode them using the NLP model. 
We observed that these datasets contain several high-frequency words (e.g., \colons{Ingredients} in RecipeQA) in both correct and incorrect choices leading to bias. 
We circumvent this issue by clustering all the titles and pick answer choices uniformly over each cluster. For each cluster, we assign a fixed budget for picking titles.  
For each correct choice we select its nearest neighbors using k nearest neighbor and use adaptive filtering to remove the very similar and distant titles. We then randomly selected the incorrect choice from these titles and decreased the budget value for
the clustered index pointed by the sampled title.
If the budget for a sampled title becomes zero, we do not sample from that cluster. 
Please refer to \autoref{table:biased} and \autoref{table:debiased} for quantitative results. Our results show that the algorithm removes bias from the datasets to prevent models from memorizing hidden patterns present in the question-answer pairs. We also removed examples where the length of the instruction steps was less than six.

\IncMargin{1.5em}
{\SetAlgoNoLine%
\begin{algorithm}
    \DontPrintSemicolon
    \SetKwData{Left}{left}
    \SetKwData{Up}{up}
    \SetKwFunction{FindCompress}{FindCompress}
    \SetKwInOut{Input}{input}
    \SetKwInOut{Output}{output}
    
\Indm\Indmm
\Input{dataset ($\mathcal{D}_{train}$, $\mathcal{D}_{val}$), trained textual encoder (Txt-E), budget $\beta$, \textit{step\_titles}, nchoices}
\Output{debiased dataset $\mathcal{D}^{\prime}$}
\Indp\Indpp
\BlankLine
embedding = $\emptyset$ \;
\For{title $\in$ step\_titles} {
    embedding = embedding $\cup$ Txt-E(\textit{title}) \;
}
knearestneighbor = kNN(embedding) \;
cluster = KMeans(embedding)\; 
budget = \{id: $\beta$\}, $\forall$ id $\in$ cluster\;
$\mathcal{D}^{\prime}$ = $\emptyset$ \;
\For{sample $\in$ $\mathcal{D}_{train}$}{
    query = Txt-E(correct\_answer from sample)\;
    idx, distance = knearestneighbor(query)\;
    mean\_distance = mean(distance) \;
    farthest\_idx = idx[distance $>$ mean\_distance] \;
    cluster\_ids = kmeans[farthest\_idx] \;
    new\_choice\_list = $\emptyset$\;
    \For{i=1 to nchoices-1}{ 
        random\_idx = random(farthest\_idx) \;
        cluster\_id = cluster\_ids[random\_idx] \;
        sampled\_title = step\_titles[random\_idx] \;
        \While{budget[cluster\_id] $\leq$ 0}{
            Go To Line 17 ;
        }
        budget[cluster\_id] -= 1 \;
        \tcp*[h]{don't add duplicate} \;
        new\_choice\_list = new\_choice\_list $\cup$ sampled\_title \;
    }
    \tcp*[h]{Insert correct answer at index in sample} \;
    new\_choice\_list = new\_choice\_list $\cup$ correct\_answer \;
}
return $\mathcal{D}^{\prime}$ \;
\caption{Removing Bias in Dataset}
\label{algorithm}
\Indp\Indpp
\end{algorithm}}%
\DecMargin{1.5em}


%% file: exp.tex
\section{Experiments}


\begin{table*}[!htbp]
    \adjustbox{max width=\textwidth}{%
    \centering
    \begin{tabular}{lccccccc}
        \hline
                                                                              & \multicolumn{3}{c}{Word2Vec}                                                                     & \multicolumn{1}{c}{} & \multicolumn{3}{c}{BERT}                                                     \\ \cline{2-8} 
        Baseline Model/Dataset                                                                 & \multicolumn{1}{l}{RecipeQA} & \multicolumn{1}{l}{DecorationQA} & \multicolumn{1}{l}{WoodworkQA} & \multicolumn{1}{l}{} & \multicolumn{1}{l}{RecipeQA} & \multicolumn{1}{l}{DecorationQA} & \multicolumn{1}{l}{WoodworkQA} \\ \hline
        1 \ Chance & $25.00$ & $25.00$ & $25.00$ & & $25.00$ & $25.00$ & $25.00$ \\
        2 \ Hasty Student (choice only) &  $40.73$   &  $45.76$ &    $48.84$   &  &  $40.94$   &  $54.76 $  &   $53.91$     \\
        3 \ Hasty Student      &        $46.77$     &        $46.12$      &       $49.69$      &      &    $47.22$      &          $54.56$  &   $53.72$         \\
        4 \ Impatient Reader (Text only)     &   $50.31$  &    $39.22$   &      $38.58$    &     &    $53.02$    &  $53.55$   &    $49.67$        \\
        \hline
    \end{tabular}}
    \\
    \caption{Results for the baseline models trained on the biased dataset of RecipeQA, DecorationQA, WoodworkQA on Word2Vec and BERT.}
    \label{table:biased}
\end{table*}

\begin{table*}[!htbp]
    \adjustbox{max width=\textwidth}{%
    \centering
    \begin{tabular}{lccccccc}
        \hline
                                                                              & \multicolumn{3}{c}{Word2Vec}                                                                     & \multicolumn{1}{c}{} & \multicolumn{3}{c}{BERT}                                                     \\ \cline{2-8} 
        Baseline Model/Dataset                                                                 & \multicolumn{1}{l}{RecipeQA} & \multicolumn{1}{l}{DecorationQA} & \multicolumn{1}{l}{WoodworkQA} & \multicolumn{1}{l}{} & \multicolumn{1}{l}{RecipeQA} & \multicolumn{1}{l}{DecorationQA} & \multicolumn{1}{l}{WoodworkQA} \\ \hline
        1 \ Chance          &    $25.00$    &    $25.00$    &           $25.00$    &     &  $25.00$    &   $25.00$     &  $25.00$ \\
        2 \ Hasty Student (choice only) &        $32.23$   &     $31.90$    &   $36.00$      &    &   $34.27$    &   $37.93$   &    $45.02$         \\
        3 \ Hasty Student    &  $32.92$    &   $32.11$    &   $36.34$    &     &     $33.33$      &     $38.36$     &   $44.98$         \\
        4 \ Impatient Reader (Text only)   &    $33.52$   &   $30.75$  &    $29.55$    &   &     $30.31$    &   $37.50$  &   $41.71$         \\
        \hline
    \end{tabular}}
    \\
    \caption{Results for the baseline models trained on the debiased dataset of RecipeQA, DecorationQA, WoodworkQA on Word2Vec and BERT.}
    \label{table:debiased}
\end{table*}

\subsection{Baseline Models} \label{baselinemodels}

We evaluate four strong baseline models on the three M3C datasets-- DecorationQA, WoodworkQA, and RecipeQA-- for textual cloze question answering task. These baselines are some of the most commonly used procedures in other M3C  works \cite{yagcioglu2018recipeqa,tapaswi2016movieqa}. For this paper, our baselines focus only on the textual part of the context. We use these baselines to provide evidence of the bias present in the data and also the efficacy of our algorithm in removing that bias. 

\textbf{Hasty Student} \cite{yagcioglu2018recipeqa} matches each choice to the question list without looking at the context. Each question and choice are first embedded using an LSTM. Each choice is then scored by computing its distance with the questions in the feature space. We also propose a variant of the above model-- \textbf{Hasty Student only Choice}, where we directly score the choices by first encoding them with an LSTM and then using a linear layer to produce logits. This model mimics a student who would try to solve comprehension tasks by only looking at the choices. The scores for each answer choice is then fed to a loss function for training.


\textbf{Impatient Reader} \cite{hermann2015teaching} is an attention-based model that recurrently uses attention over the context for each question except the location containing the \colons{$@$ placeholder}. This attention allows the model to accumulate information recurrently from the context as it sees each question embeddings. It outputs a final joint embedding for the answer prediction. This embedding is used to compute a compatibility score for each choice using a cosine similarity function in the feature space. The attention over context and question is computed on the output of an LSTM. The answer choices are also encoder using an LSTM with a similar architecture. 




\subsection{Implementation Details}
We used Word2Vec \cite{mikolov2013efficient} and BERT \cite{devlinetal2019bert} to represent the word level embeddings on the textual context, question, and choices. We prepared our Word2Vec model using all the step titles in the datasets. We set the embedding size for Word2Vec as 100$-$d. For BERT, we used BERT base model (uncased) for Sentence Embeddings i.e., \colons{bert-base-nli-mean-tokens} pre-trained model from SentenceTransformers \cite{reimers-2019-sentence-bert}. 
For the baseline models, we use a stacked bidirectional LSTM with three layers and a dropout of 0.9. The steps mentioned above are consistent with all our experiments to create embeddings for context, question, and choices. We used Adam \cite{kingma2015adam} optimizer and fixed the learning rate to $5e-4$ with early stopping criteria with a patience set of $20$ for all of our experiments. We performed our experiments on a single GTX 1080Ti that took about $8-10$ hours for a single experiment. We employed the same hyperparameters for all the baseline models.

\subsection{Results}
We begin by verifying the bias present in the data via the baseline models in \autoref{table:biased}. The table presents results with the two textual embeddings, i.e., Word2Vec (left) and BERT (right). The naive baseline \textbf{Hasty Student only Choice} that only uses the answer choices to produce a score improves performance over the baseline by a range of $15$\%-$30$\% (row 2 - row 3).  
We observe a similar behavior with the BERT-based textual embeddings, which corroborates our findings of the bias in the dataset. 
We show the results on the debiased data in \autoref{table:debiased} where we observe a considerable drop in the performances across all the baselines by $8-17$\%.
Our claim is also supported by a similar drop in the naive baseline model. 
This decrease in accuracy suggests that the debiased algorithm has distributed the step titles equally from the clusters as alternate choices for the questions. Thus the trained model can no longer benefit from the hidden artifacts present in the original dataset due to the generation method proposed in \cite{yagcioglu2018recipeqa}.

We further observe that simple attention-based approaches such as the ImpatientReader, are also not able to improve accuracy further. The above statement holds for both the biased and the debiased datasets.
The results in \autoref{table:debiased} is closer to the chance performance compared to \autoref{table:biased}, caused due to the removal of the bias and also the inability of these baseline models to take advantage of the information present in the context.
The above results suggest that the procedural context in our M3C datasets needs a more efficient solution that can comprehend (1) the temporal and causal factors, and (2) exploit the multimodality in the inputs. 
Another challenging aspect of this dataset is that these datasets are obtained from a publicly available website that has collected DIY projects from a wide variety of anonymous users in an unconstrained environment. In such cases, it is difficult to eliminate inconsistency, missing information, and biases present in the data. 
Our results suggest that the proposed \autoref{algorithm} has removed possible bias introduced due to the choice generation method described in \cite{yagcioglu2018recipeqa}. Finally, our results with BERT seem to be close to the Word2Vec model and in some cases has improved by a $10$\% margin.


%% file: conclusion.tex
\section{Conclusion}
\label{sec:conclusion}

This paper focuses on the multimodal machine comprehension question answering task, especially where the context is procedural. To this end, we introduced two new large datasets WoodworkQA, DecorationQA as a testbed for evaluation of procedural M3C tasks. WoodworkQA consists of 16K wood-crafting procedures; similarly, DecorationQA consists of 10K decorating projects. Our current work addresses only the textual cloze style questions, mainly using the natural language to expose bias in naturally occurring datasets. During our study into procedural M3C task, we discovered bias in the question-answer preparation procedure proposed by an earlier work \cite{yagcioglu2018recipeqa}. We thus proposed an automated method (\autoref{algorithm}) capable of modifying the given dataset to remove the bias elements.

We hope that our work and evaluation will encourage further research in this area. We will also release the new splits and the datasets for the community. In addition, we would like to note several research directions to develop
this area further.

\begin{enumerate}
    \item We intend to extend these datasets to allow evaluation for the remaining
    three cloze tasks-- \colons{visual cloze}, \colons{visual coherence}, \colons{visual ordering}  
    \item We also plan on building algorithms that can effectively exploit
    the multimodality in the input context. Multimodality
    is essential in many real-world cases as users tend to
    often describe information in the modality that is convenient
    for that step.    
    \item Build models to understand better the temporal and
    causal factors in the given context. For example, it
    would be interesting to model a
    particular object's state changes during the preparation phase.
    \item Exploring the usefulness of large pre-trained (or self-supervised) models from both NLP domain and visual domain \cite{miech2020end}. 
\end{enumerate}

%

